\pdfoutput=1

\documentclass[11pt]{article}

\usepackage{acl}

\usepackage{times}
\usepackage{latexsym}

\usepackage{booktabs}
\usepackage{multirow}

\usepackage{tcolorbox}
\usepackage{xcolor}
\definecolor{bggray}{RGB}{245,245,245} 

\usepackage[T1]{fontenc}

\usepackage[utf8]{inputenc}

\usepackage{microtype}

\usepackage{inconsolata}

\usepackage{graphicx}
\usepackage{placeins} 
\usepackage{needspace} 
\usepackage{dblfloatfix} 
\usepackage{amsmath}
\usepackage{mathrsfs}
\usepackage{enumitem}

\usepackage{algorithm} 
\usepackage{algorithmic}
\usepackage{float}

\usepackage{pifont}
\newcommand{\circnum}[1]{\ding{\numexpr171+#1}}

\usepackage{tabularx}   
\usepackage{multirow}   
\usepackage{array}      
\usepackage{booktabs}   
\usepackage{enumitem}   

\usepackage{tcolorbox}
\usepackage{caption}

\tcbset{colback=white, colframe=black, fonttitle=\bfseries}

\title{Orion-RAG: Path-Aligned Hybrid Retrieval for Graphless Data }

\author{
Zhen Chen\textsuperscript{1}\thanks{Equal contribution.}\quad
Weihao Xie\textsuperscript{1}\footnotemark[1]\quad
Peilin Chen\textsuperscript{1}\quad
Shiqi Wang\textsuperscript{1}\quad
Jianping Wang\textsuperscript{1}\\
\textsuperscript{1}City University of Hong Kong, Hong Kong SAR \quad
\\
\texttt{\{zchen979-c@my., weihaxie-c@my., plchen3@, shiqwang@, jianwang@\}cityu.edu.hk}
}

\newcommand{\tx}[1]{{\textcolor{blue}{#1 \#cxt}}}

\begin{document}
\maketitle
\begin{abstract}
Retrieval-Augmented Generation (RAG) has proven effective for knowledge synthesis, yet it encounters significant challenges in practical scenarios where data is inherently discrete and fragmented. 
In most environments, information is distributed across isolated files like reports and logs that lack explicit links. 
Standard search engines process files independently, ignoring the connections between them. Furthermore, manually building Knowledge Graphs is impractical for such vast data. To bridge this gap, we present Orion-RAG. Our core insight is simple yet effective: we do not need heavy algorithms to organize this data. Instead, we use a low-complexity strategy to extract lightweight ``paths'' that naturally link related concepts. 
We demonstrate that this streamlined approach suffices to transform fragmented documents into semi-structured data, enabling the system to link information across different files effectively. Extensive experiments demonstrate that Orion-RAG consistently outperforms mainstream frameworks across diverse domains, supporting real-time updates and explicit Human-in-the-Loop verification with high cost-efficiency. 
Experiments on FinanceBench demonstrate superior precision with a 25.2\% relative improvement over strong baselines.
\end{abstract}



\section{Introduction}

Retrieval-Augmented Generation (RAG) \citep{lewis2020retrieval} integrates retrieval mechanisms \citep{salton1983introduction} with large language models (LLMs) to enhance generation using external data. By combining parametric knowledge with external evidence, RAG has become essential in knowledge-intensive domains, such as healthcare \citep{medrag2024}, legal compliance \citep{legalrag2024}, finance \citep{financebench_ref}, enterprise support \citep{customerrag2024}, and scientific workflows \citep{scirag2024}. RAG systems generally offer better accuracy and document understanding compared to standalone LLMs.

However, deploying RAG in real-world scenarios entails challenges extending beyond accuracy, specifically the enterprise demands for rapid deployment and operational controllability. Fundamentally, real-world data is fragmented: it consists of discrete text units with no explicit links connecting them. For example, when a user queries a specific company's revenue, standard retrievers may fail to locate the relevant financial statement. This occurs because the statement contains only numerical data without explicitly repeating the company name, resulting in zero lexical overlap with the query. Beyond fragmentation, a critical barrier is the absence of pre-constructed graphs in real-world data, which forces heavy Knowledge Graph (KG) approaches \citep{edge2024graphrag, yasunaga2021qagnn, pan2024unifying} to undertake computationally expensive global construction. Furthermore, frequent updates from multiple users make maintaining these rigid structures impractical, as re-indexing limits real-time concurrency. 
Lastly, ``black-box'' retrieval lacks the transparency required for Human-in-the-Loop (HITL) verification, which is critical for industrial adoption.

\begin{figure*}[!t]
    \centering
    \vspace{-4mm}
    \makebox[\textwidth][c]{\includegraphics[width=1\textwidth,page=1,trim=0 15 0 15,clip]{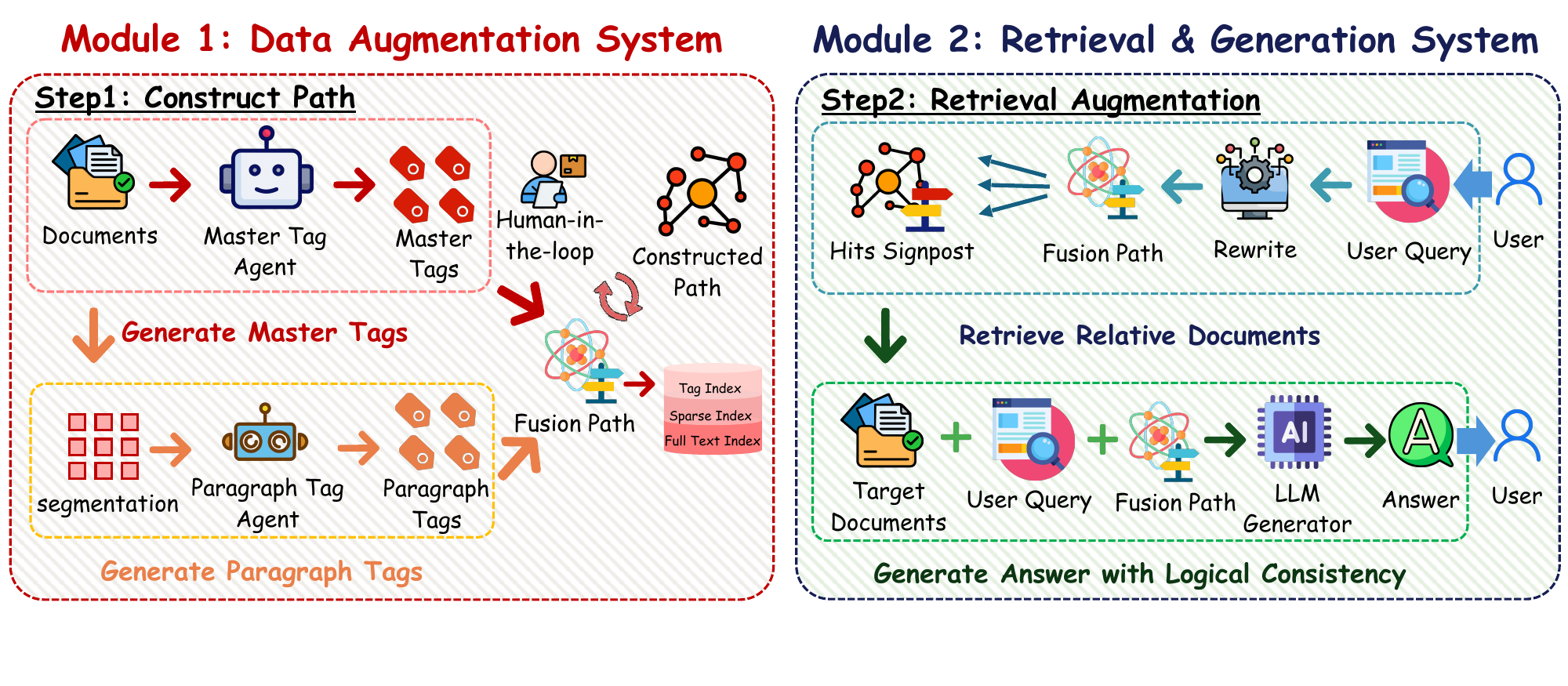}}
    \caption{System overview of Orion-RAG. (1) The Path-Annotation Data Augmentation subsystem (left) employs dual-layer labeling agents to construct hierarchical navigation paths from fragmented text, enabling real-time incremental indexing. (2) The Multi-Layer Hybrid Retrieval subsystem (right) utilizes these paths as explicit logical signposts, integrating sparse and dense search to guide the generator towards accurate and interpretable answers.}
    \label{fig:system}
    \vspace{-4mm}
\end{figure*}

To address these challenges, we propose Orion-RAG, a hybrid retrieval framework designed for agile and lightweight implementation.
It uncovers latent structures without the overhead of pre-built KGs or complex global modeling. Our system consists of two main modules:

\noindent \textbf{Path-Annotation Data Augmentation} (see Fig.~\ref{fig:system} left): This module generates a Path, which is a hierarchical list of textual keyword tags created by our dual-layer labeling system.
This path acts as a navigational ``map'' for retrieval, where each tag serves as a ``signpost''. By embedding the path into a single averaged vector, a query matching any single signpost significantly boosts the recall of the target document. Beyond retrieval, these paths can be injected as structural prompts to the generator, enhancing the logical consistency of the answer. Crucially, this process operates locally on data segments, allowing for real-time incremental updates as new documents are uploaded by multiple users. The generated ``ephemeral knowledge graph'' provides explicit reasoning paths, making the system interpretable and supporting human-in-the-loop auditing.

\noindent \textbf{Multi-Layer Hybrid Retrieval} (see Fig.~\ref{fig:system} right): This module integrates sparse retrieval and dense semantic search with path-based indexing.
It utilizes the paths as explicit logical chains of evidence. A sophisticated search algorithm is then employed to rigorously optimize retrieval accuracy while maintaining high computational efficiency.

Orion-RAG is explicitly engineered for industrial scalability. Unlike methods that require complex recursive processing or global graph clustering, we generate connection paths following a linear design. This algorithmic simplicity ensures low computational complexity, making the system concurrency-friendly and cost-effective. 

The evaluation spans three diverse datasets: Mini-Wiki (general knowledge) \citep{mini_wiki_ref}, FinanceBench (financial reports) \citep{financebench_ref}, and SeaCompany, a custom-curated corpus of Southeast Asian company profiles designed to test retrieval on highly fragmented data. To simulate realistic ``graphless'' environments, our experimental setup excludes the use of pre-existing KGs. Performance is benchmarked against a comprehensive suite of strong baselines, encompassing widely adopted sparse, dense, and hybrid RAG methods.
The results show that Orion-RAG achieves superior performance in Hit Rate, Precision, BERTScore , and ROUGE-L. 
Moreover, the proposed method operates with {linear complexity $\mathcal{O}(N)$}, ensuring that even real-time incremental updates remain computationally lightweight and scalable.


We propose Orion-RAG, a framework that resolves complex data fragmentation using lightweight path structures. This design eliminates the dependency on heavy KGs, ensuring high-concurrency scalability without compromising retrieval depth. Furthermore, we introduce a linear, low-complexity augmentation module that enables real-time and multi-user updates while generating explicit paths for human-in-the-loop verification. To validate these capabilities, we present SeaCompany, a challenge benchmark of Southeast Asian corporate profiles and QA pairs characterized by high data fragmentation, serving as a specialized testbed for evaluating retrieval integration. Finally, experiments on Mini-Wiki, FinanceBench, and SeaCompany demonstrate that Orion-RAG outperforms existing baselines in accuracy and generation quality, providing a practical, cost-efficient solution for processing discrete and fragmented data with LLMs.
\section{Related Work}

\noindent \textbf{Retrieval-Augmented Generation.}
Retrieval-Augmented Generation (RAG) integrates information retrieval with large language models (LLMs) to ground responses in external corpora, reducing hallucinations and improving accuracy. Typical pipelines comprise document chunking, embedding (e.g., Sentence-BERT \cite{reimers2019sentence}, BERT \citep{devlin2019bert} or OpenAI Ada \citep{RAG-Embedding:ADA}), retrieval via sparse methods such as BM25 \citep{RAG-Retrieve:BM25}, dense methods (FAISS \citep{RAG-VDB:FAISS}, DPR \citep{RAG-VDB:DPR}), or hybrid rank fusion (e.g., RRF \citep{cormack2009reciprocal}), optional reranking (MonoT5 \citep{RAG-Rank:MonoT5}, ColBERT \citep{RAG-Rank:Colbert}), and generation conditioned on retrieved context. Evaluation commonly reports EM, F1, BLEU for generation, and Recall@k, MRR, nDCG for retrieval.

\noindent \textbf{Reasoning-Action Interleaving.} ReAct \citep{react_ref} interleaves chain-of-thought \citep{wei2022chain} ``reasoning traces'' with task-specific actions, enabling LLMs to plan, query tools (e.g., a Wikipedia API), and update decisions with external feedback. Empirically, ReAct improves multi-hop QA (HotpotQA) and fact verification (FEVER) over baselines without explicit reasoning or action interleaving, and outperforms imitation/RL agents on ALFWorld and WebShop with few-shot prompting. This line of work highlights that retrieval and tool use can be guided by explicit intermediate reasoning to mitigate hallucinations and expose interpretable trajectories.

\noindent \textbf{Agentic Information Sieving.}  DeepSieve \citep{deepsieve_ref} adopts an agentic architecture to facilitate precise information sieving through an LLM-based knowledge router. The framework decomposes multifaceted queries into structured sub-questions, which are then recursively routed to the most relevant knowledge sources to filter noise via multi-stage distillation. While this approach enhances retrieval accuracy, its dependence on sequential LLM calls for routing and filtering introduces significant overhead in terms of latency and computational cost. Though DeepSieve enhances reasoning depth and retrieval precision by leveraging modular agents to reduce noise, its reliance on iterative LLM calls for routing and filtering can introduce significant latency and token costs, posing challenges for high-throughput real-time applications.

\noindent \textbf{Recursive Context Aggregation.} RAPTOR \citep{raptor_ref} advances long-context retrieval by recursively clustering and summarizing text chunks into a hierarchical tree. While this effectively captures multi-level abstractions, the reliance on global clustering algorithms introduces inherent challenges for dynamic data environments. Primary among these is the difficulty of incremental updates: since higher-level summaries are derived from specific clusters of lower-level nodes, injecting new documents often necessitates re-clustering and re-summarizing large portions of the tree structure. Consequently, this creates a significant dependency on the global dataset state, resulting in higher computational complexity and maintenance overhead compared to methods that process documents independently.

\noindent \textbf{Hybrid Retrieval over Semi-Structured KBs.}  HybGRAG~\citep{hybgrag_ref} studies ``hybrid'' questions over semi-structured knowledge bases (SKBs) that require both textual and relational evidence. It introduces a modular retriever bank (text versus hybrid graph+text) and a critic module for self-reflective routing refinement, showing strong gains on STaRK benchmarks. While effective in SKB settings, the approach assumes access to a pre-built knowledge graph and focuses on routing between semantic retrieval and graph retrieval.

\smallskip
Recent advancements, such as RAPTOR and agentic frameworks, mark a significant transition from flat retrieval to structure-aware reasoning. However, these methods often rely on computationally intensive recursive processes or complex graph prerequisites. Orion-RAG addresses this trade-off by introducing a lightweight, linear alternative. By generating interpretable path structures in a single pass, our approach captures the benefits of hierarchical context while eliminating the overhead of complex graph construction, offering a streamlined and scalable solution for dynamic real-world data.

\section{Methodology}
\label{sec:methodology}

The Orion-RAG architecture comprises two integrated subsystems: an offline Data Augmentation System for graphless structure induction, and an online Retrieval and Generation System for context-aware inference.

\subsection{Data Augmentation System}
\label{subsec:augmentation}

To address the challenge of navigating unstructured data without explicit graph schemas, we propose an automated Data Augmentation System that induces latent structure within a raw corpus $\mathcal{D}$. Formally, the system transforms each document $d \in \mathcal{D}$ into a set of augmented chunks $\mathcal{C}_{aug} = \{(c_i, P_i)\}$, where $c_i$ represents the textual content and $P_i$ denotes a synthesized \textit{Semantic Path}. The pipeline comprises four sequential modules:

\noindent \textbf{Master Tag Generation.}
To prevent context loss, the system first extracts global attributes from the parent document. An LLM function $f_{\text{master}}$ autonomously induces high-level descriptors $T_m$ (e.g., \textit{Entity: Tesla, Type: 10-K Report}). These tags are then attached to every subsequent text chunk, ensuring that even isolated fragments retain their global context (e.g., distinguishing generic ``Revenue'' figures across different years).

\noindent \textbf{Document Segmentation.}
The document $d$ is then divided into a sequence of micro-chunks $\mathcal{C} = \{c_1, \dots, c_k\}$ using a context-aware sliding window. This ensures each $c_i$ retains sufficient local information for independent analysis.

\noindent \textbf{Paragraph Tag Generation.}
To capture fine-grained semantics, we define a function $f_{\text{para}}$ that generates local descriptors $T_p^{(i)} = f_{\text{para}}(c_i)$. This functions as a bottom-up implicit clustering mechanism: semantically similar chunks across disjoint documents independently yield similar $T_p$ values, logically connecting disparate information islands without global clustering algorithms.

\noindent \textbf{Path Construction.}
We synthesize the final structure by fusing global and local contexts. For each chunk $c_i$, we construct a hierarchical Semantic Path:
\begin{equation}
    P_i = T_m \oplus T_p^{(i)}
\end{equation}
This results in a structured coordinate (e.g., $[\textit{Project X} \rightarrow \textit{Finance} \rightarrow \textit{Q3 Budget}]$). A Human-in-the-Loop checkpoint allows experts to verify these readable tags before indexing.

\noindent \textbf{Hybrid Index Construction.}
Finally, we encode the augmented chunks into three parallel indices to support multi-dimensional retrieval:
\begin{itemize}[nosep]
    \item \textbf{Tag Index ($\mathcal{I}_{\text{tag}}$):} Dense vectors $\mathbf{v}_{\text{path}} = \text{Embed}(P_i)$ capturing structural logic.
    \item \textbf{Full-Text Index ($\mathcal{I}_{\text{dense}}$):} Dense vectors $\mathbf{v}_{\text{text}} = \text{Embed}(c_i)$ capturing semantic meaning.
    \item \textbf{Sparse Index ($\mathcal{I}_{\text{sparse}}$):} An inverted index of $c_i$ (BM25) for exact lexical matching.
\end{itemize}

\subsection{Retrieval and Generation System}
\label{subsec:rag_system}

This subsystem functions as a mapping function $f_{RAG}: q \rightarrow A$, transforming a potentially ambiguous user query $q$ into a grounded answer $A$. The pipeline consists of five sequential stages designed to maximize the signal-to-noise ratio at each step.

\noindent \textbf{Query Rewriting.}
To mitigate the semantic gap between user intent and index representation, we first employ a Query Rewriting Agent. Given a raw query $q$, the agent generates a set of optimized sub-queries $Q'$:
\begin{equation}
    Q' = \{q\} \cup \{q'_1, \dots, q'_n\}
\end{equation}

where each $q'_i$ targets a specific latent aspect (e.g., expanding explicit entities or resolving acronyms). This expansion ensures comprehensive coverage across both semantic and lexical search spaces, similar to hypothetical document embedding strategies \cite{gao2022precise}.


\noindent \textbf{Coarse Retrieval.}
For each generated sub-query $q'_k \in Q'$, we execute an independent retrieval cycle to prioritize recall. We perform parallel searches on the constructed indices:
\begin{itemize}[nosep]
    \item \textbf{Path-Based Retrieval:} Queries the Tag Index ($\mathcal{I}_{\text{tag}}$) to find chunks whose structural paths align with the sub-query's intent. This effectively filters the search space using the induced document structure.
    \item \textbf{Lexical Retrieval:} Queries the Sparse Index ($\mathcal{I}_{\text{sparse}}$) to capture exact keyword matches that dense vectors might miss.
\end{itemize}
The union of these results constitutes the candidate set of augmented pairs $\mathcal{C}_{cand} = \{(c, P)\}$ specific to the current sub-query $q'_k$, ensuring the semantic path travels with the content.

\noindent \textbf{Weighted Re-ranking.}
To refine the candidate set $\mathcal{C}_{cand}$ and prioritize precision, this module selects the optimal top-$k$ context for the specific sub-query. We employ a Weighted Reciprocal Rank Fusion (RRF) algorithm \citep{cormack2009reciprocal} that integrates scores from three perspectives: structural ($R_{\text{tag}}$), semantic ($R_{\text{sem}}$), and lexical ($R_{\text{sparse}}$). The score is calculated relative to the sub-query $q'_k$:
\begin{equation}
    S_{\text{RRF}}(c, q'_k) = \sum_{j \in \{\text{tag, sem, sparse}\}} \frac{w_j}{\eta + R_j(c, q'_k)}
\end{equation}
where $R_j(c, q'_k)$ is the rank of the chunk in that specific index regarding $q'_k$. The top-$k$ chunks ($\mathcal{C}_{top}$) with the highest scores are passed to the next stage.


\noindent \textbf{Document Pruning.}
Given the multiplicity of sub-queries, directly aggregating all retrieved contexts would overwhelm the generator's context window and degrade performance due to relevant information being lost in noise \cite{liu2023lost} \cite{shi2023large}. To prevent this, a Pruning Agent performs granular filtering on the retrieved chunks. For each sub-query $q'_k$, the agent analyzes its top candidates $(c, P)$ via a filtering function $\phi$:

\begin{equation}
    \mathcal{C}_k = \{ \phi((c, P), q'_k) \mid (c, P) \in \mathcal{C}_{top}^{(k)} \} \setminus \{\emptyset\}
\end{equation}
Here, the explicit path $P$ (concatenated with $c$) serves as a context anchor, allowing the agent to discern and discard irrelevant chunks (e.g., matching "Revenue" but from the wrong "Year") before they reach the generator. We preserve the mapping between each sub-query and its verified evidence.


\noindent \textbf{Answer Generation.}
Finally, the Generator synthesizes the answer $A$ through Structural Context Injection. The LLM receives the original query $q$ along with the structured pairs of sub-queries and their corresponding contexts:
\begin{equation}
    A = \text{LLM}\left(Prompt\left(q, \left\{ (q'_k, \mathcal{C}_k) \right\}_{k=1}^{|Q'|} \right)\right)
\end{equation}
In the prompt, chunks in $\mathcal{C}_k$ are grouped under their respective sub-query $q'_k$. The path information retained within the pruned text serves as a structural guide, helping the LLM distinguish between similar data points from different contexts and enabling precise, explainable citations.

\subsection{Overall Algorithm}
\label{subsec:overall_algorithm}

The complete workflow of Orion-RAG is formally summarized in Algorithm \ref{alg:orion_flow}. The process integrates offline structure induction with a precise, sub-query specific online inference pipeline.

Uniquely, our online phase processes each generated sub-query independently through retrieval, re-ranking, and pruning. This ensures that context relevance is evaluated against specific information needs ($q'_k$) rather than the potentially ambiguous original query, before being aggregated for the final generation.

\begin{algorithm}[t]
\caption{Orion-RAG: Structure Induction and Inference}
\label{alg:orion_flow}

\renewcommand{\algorithmicrequire}{\textbf{Input:}}
\renewcommand{\algorithmicensure}{\textbf{Output:}}

\newcommand{\algocomment}[1]{\hfill \textcolor{blue}{\itshape /* #1 */}}
\newcommand{\phaseheader}[1]{\vspace{0.2cm} \STATE \textbf{\textcolor{blue}{#1}}}

\begin{algorithmic}[1]
\REQUIRE Raw Corpus $\mathcal{D}$, User Query $q$
\ENSURE Grounded Answer $A$

\phaseheader{/* Phase 1: Offline Structure Induction */}
\STATE Initialize Indices $\mathcal{I} = \{\mathcal{I}_{tag}, \mathcal{I}_{dense}, \mathcal{I}_{sparse}\}$
\FOR{each document $d \in \mathcal{D}$}
    \STATE $\mathcal{I} \leftarrow \text{DataAugmentationSystem}(d, \mathcal{I})$ 
\ENDFOR

\phaseheader{/* Phase 2: Online Inference */}
\STATE $Q' \leftarrow \{q\} \cup \text{Rewriter}(q)$ 
\STATE $\mathcal{S}_{ctx} \leftarrow \emptyset$ 

\FOR{each sub-query $q'_k \in Q'$}
    \STATE \textbf{// Stage 1: Coarse Retrieval}
    \STATE $\mathcal{C}_{cand} \leftarrow \text{HybridRetriever}(q'_k, \mathcal{I})$ \algocomment{Returns pairs $(c, P)$}

    \STATE \textbf{// Stage 2: Weighted Re-ranking}
    \STATE $\mathcal{C}_{top} \leftarrow \text{WeightedRanker}(\mathcal{C}_{cand}, q'_k)$

    \STATE \textbf{// Document Pruning}
    \STATE $\mathcal{C}_k \leftarrow \text{Pruner}(\mathcal{C}_{top}, q'_k)$ 

    \STATE $\mathcal{S}_{ctx} \leftarrow \mathcal{S}_{ctx} \cup \{(q'_k, \mathcal{C}_k)\}$ 
\ENDFOR

\phaseheader{/* Generation with Structure Injection */}
\STATE $A \leftarrow \text{Generator}(q, \mathcal{S}_{ctx})$

\RETURN $A$
\end{algorithmic}
\end{algorithm}

\section{Experiments}
\label{sec:experiments}

To rigorously evaluate the efficacy of Orion-RAG in graphless, semi-structured environments, we conducted experiments across three distinct benchmarks. FinanceBench \citep{financebench_ref} serves as a long-document testbed mimicking enterprise-scale retrieval, while Mini-Wiki \citep{mini_wiki_ref} assesses general knowledge through synthetic documents. Additionally, we introduce SeaCompany, a manually constructed dataset of Southeast Asian corporate profiles generated via RAGen \citep{tian2025ragen}, designed to test retrieval on highly fragmented information islands. Detailed statistics and pre-processing steps are provided in Appendix \ref{sec:appendix_benchmarks}.

We compare Orion-RAG against seven strong baselines categorized into Naive RAG (Dense, Sparse, Hybrid, Re-ranking) and Agentic/Iterative RAG (ReAct \citep{react_ref}, DeepSieve \citep{deepsieve_ref}, RAPTOR \citep{raptor_ref}). These methods represent diverse paradigms ranging from standard retrieval to advanced recursive summarization and iterative reasoning. To ensure a strictly fair comparison, all baselines were standardized regarding embedding models and generator architectures. Specific implementation details are provided in Appendix \ref{subsubsec:orion_impl} and \ref{subsubsec:baselines}.



We comprehensively evaluated both generation quality and retrieval precision. For generation performance across FinanceBench, Mini-Wiki, and SeaCompany, we report ROUGE-L~\citep{rouge_ref} to measure structural fidelity via longest common subsequences, and BERTScore (F1)~\citep{bertscore_ref} to evaluate semantic similarity. Unlike simple lexical overlap, BERTScore utilizes contextual embeddings to assess meaning preservation, aligning more closely with human judgment in open-ended tasks.

Retrieval evaluation focuses on FinanceBench and SeaCompany, excluding Mini-Wiki due to synthetic ambiguity. Given our data augmentation approach, we rigorously define the ground truth set $G_q$ for a query $q$ as all sub-chunks derived from the specific source document associated with that query. Based on this, we report two metrics across $k \in \{3, 5, 10\}$. Hit Rate@k measures the system's ability to locate at least one correct information boundary. Crucially, we also evaluate Precision, defined as the proportion of relevant chunks within the retrieved set ($|Retrieved_k \cap G_q| / k$). This metric is vital for validating our multi-path strategy; while high recall is desirable, low precision implies a high density of irrelevant ``distractor'' documents, which introduces noise and significantly increases the risk of hallucination in the downstream generator.


\subsection{Retrieval Evaluation}
\label{subsec:retrieval_perf}

This section evaluates the core competency of the Orion-RAG retrieval module: its ability to accurately isolate relevant information from large-scale, unstructured corpora. We aim to demonstrate that a structural, path-based index can achieve high recall without the precision degradation typically associated with naive query expansion or iterative retrieval methods.

\begin{figure*}[!t]
    \centering
    \includegraphics[width=\textwidth]{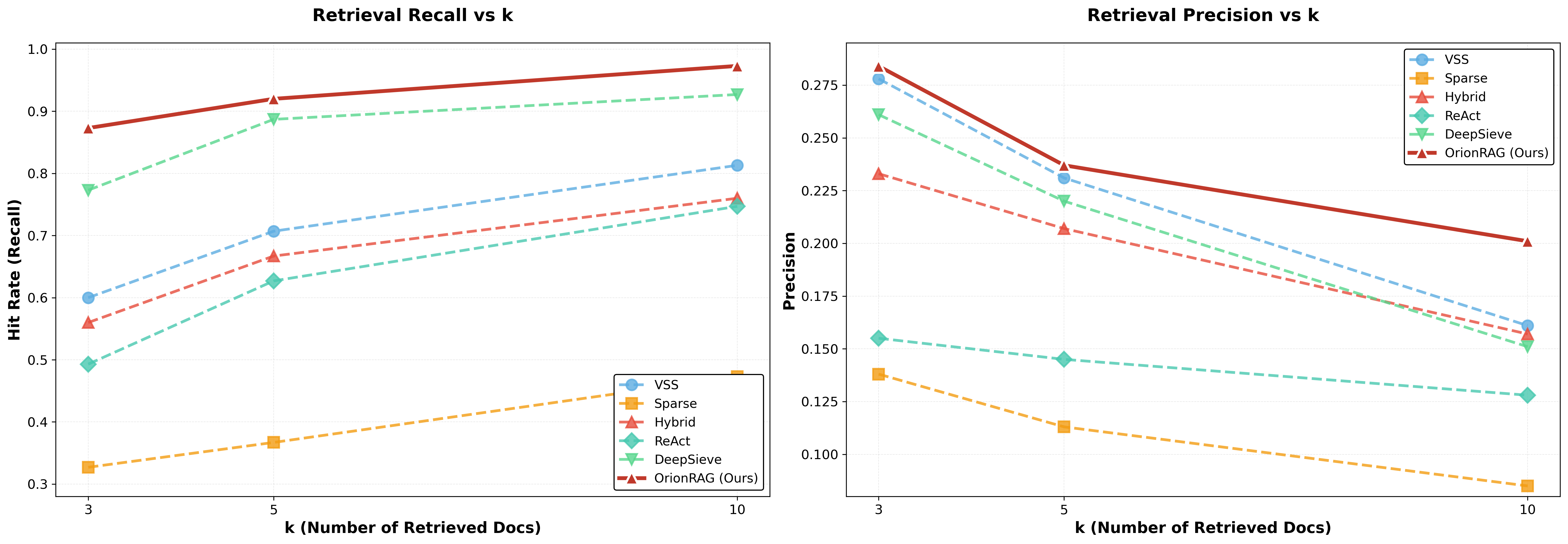}
    \caption{Retrieval Performance Comparison. Orion-RAG demonstrates a superior balance of Hit Rate and Precision across diverse datasets.}
    \label{fig:retrieval_res}
\end{figure*}

\noindent \textbf{Setup.}  We focus exclusively on FinanceBench and SeaCompany due to their explicit chunk-level ground truth mappings, enabling precise measurement of Hit Rate@k (recall) and Precision (signal-to-noise ratio) across $k \in \{3, 5, 10\}$. Detailed prompt templates used for baseline reproductions and our system are provided in Appendix \ref{sec:prompts}. The comparative results are summarized in Table \ref{tab:retrieval_comparison_merged} and visualized in Figure \ref{fig:retrieval_res}.

\begin{table*}[!t]
    \centering
    \scriptsize 

    \caption{Retrieval Performance Comparison across FinanceBench and SeaCompany. Best results are \textbf{bolded}, and best baseline results are \underline{underlined}.}
    \label{tab:retrieval_comparison_merged}
    
    \begin{tabular*}{\textwidth}{@{\extracolsep{\fill}}l cccccc cccccc}
        \toprule
        & \multicolumn{6}{c}{\textbf{FinanceBench}} & \multicolumn{6}{c}{\textbf{SeaCompany}} \\
        \cmidrule(lr){2-7} \cmidrule(lr){8-13} 
        & \multicolumn{2}{c}{$k=3$} & \multicolumn{2}{c}{$k=5$} & \multicolumn{2}{c}{$k=10$} 
        & \multicolumn{2}{c}{$k=3$} & \multicolumn{2}{c}{$k=5$} & \multicolumn{2}{c}{$k=10$} \\
        \cmidrule(lr){2-3} \cmidrule(lr){4-5} \cmidrule(lr){6-7} 
        \cmidrule(lr){8-9} \cmidrule(lr){10-11} \cmidrule(lr){12-13}
        
        \textbf{Method} & Hit & Prec. & Hit & Prec. & Hit & Prec. & Hit & Prec. & Hit & Prec. & Hit & Prec. \\
        \midrule
        
        VSS & 0.600 & \underline{0.278} & 0.707 & \underline{0.231} & 0.813 & \underline{0.161} & 0.770 & 0.453 & 0.840 & 0.375 & 0.885 & 0.259 \\
        Sparse & 0.327 & 0.138 & 0.367 & 0.113 & 0.473 & 0.085 & 0.930 & 0.542 & 0.970 & 0.397 & 0.990 & 0.243 \\
        Hybrid & 0.560 & 0.233 & 0.667 & 0.207 & 0.760 & 0.157 & 0.955 & \underline{0.622} & \underline{0.985} & \underline{0.495} & \underline{\textbf{0.995}} & \underline{0.334} \\
        \midrule 
        
        ReAct & 0.493 & 0.155 & 0.627 & 0.145 & 0.747 & 0.128 & 0.680 & 0.374 & 0.785 & 0.339 & 0.845 & 0.232 \\
        DeepSieve & \underline{0.773} & 0.261 & \underline{0.887} & 0.220 & \underline{0.927} & 0.151 & \underline{0.965} & 0.513 & 0.980 & 0.462 & 0.985 & 0.318 \\
        \midrule
        
        \textbf{Orion-RAG (Ours)} & \textbf{0.873} & \textbf{0.284} & \textbf{0.920} & \textbf{0.237} & \textbf{0.973} & \textbf{0.201} & 0.955 & \textbf{0.727} & 0.970 & \textbf{0.559} & \textbf{0.995} & \textbf{0.342} \\
        \midrule 
        
        \textit{Rel. Improv.} & \textit{+12.9\%} & \textit{+2.3\%} & \textit{+3.8\%} & \textit{+2.9\%} & \textit{+5.0\%} & \textit{+25.2\%} & \textit{-1.0\%} & \textit{+16.9\%} & \textit{-1.5\%} & \textit{+12.9\%} & \textit{0.0\%} & \textit{+2.3\%} \\
        \bottomrule
    \end{tabular*}
\end{table*}

\noindent \textbf{Results.}  Table \ref{tab:retrieval_comparison_merged} confirms that Orion-RAG significantly outperforms baselines in both recall and precision. On FinanceBench, our method dominates with a 97.3\% Hit Rate at $k=10$ while achieving a remarkable 25.2\% relative improvement in Precision compared to the strongest baseline. Similarly, on the fragmented SeaCompany dataset, Orion-RAG excels in noise filtration, securing a 16.9\% precision gain at $k=3$ over Hybrid Retrieval. These results validate that our hierarchical tagging mechanism effectively isolates precise information without the noise accumulation typical of iterative or wide-net approaches.

\noindent \textbf{Impact of Chunk Size.} We further investigated how text segmentation granularity affects retrieval. Our ablation study indicates that a chunk size of 500 characters offers the optimal trade-off between semantic context and signal noise. Excessive context (e.g., 2000+ chars) was found to degrade precision significantly. Detailed ablation results and analysis are provided in Appendix \ref{subsec:ablation_retrieval}.



\subsection{Generation Evaluation}
\label{subsec:gen_perf}

To assess the final quality of the generated answers, we evaluated Orion-RAG against all baselines across FinanceBench, Mini-Wiki, and SeaCompany. We utilized BERTScore (F1) to measure semantic consistency and ROUGE-L to quantify structural fidelity and phrasing alignment with ground truth references.

\begin{figure*}[!t]
    \centering
    \includegraphics[width=\textwidth]{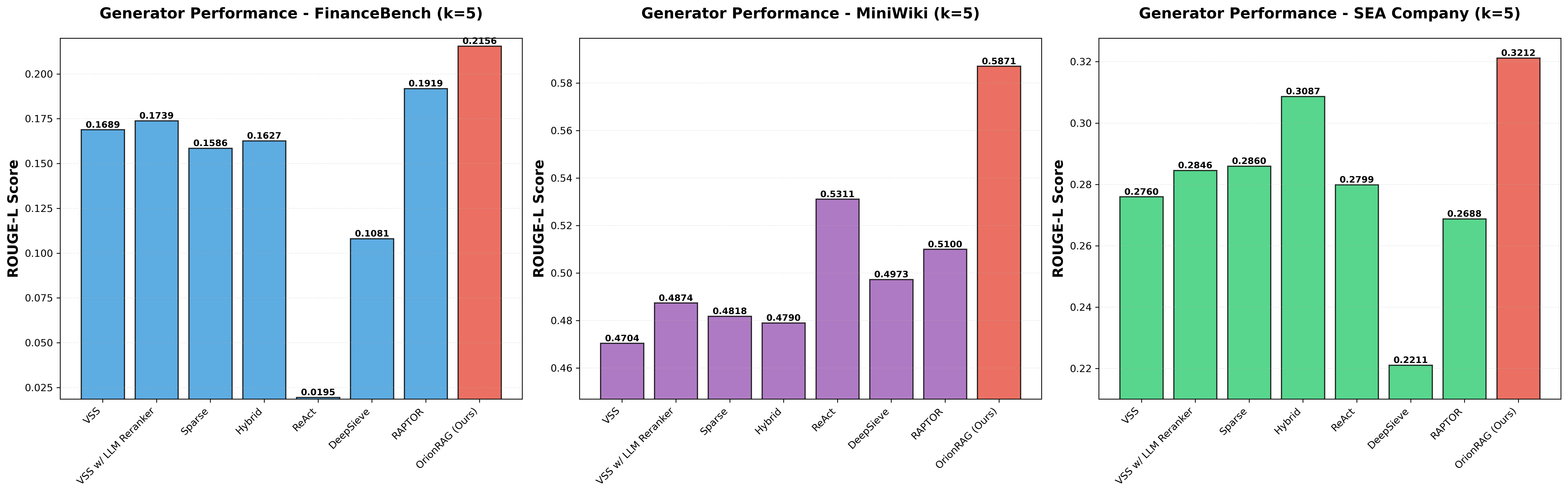}
    \caption{Generation Performance Comparison. Orion-RAG demonstrates superior semantic alignment and factual accuracy.}
    \label{fig:generation_res}
\end{figure*}

\noindent \textbf{Setup.} We compared the system's generated answers against standard ground truth references across all benchmarks. Performance was measured using \textbf{ROUGE-L} \citep{rouge_ref} to evaluate structural fidelity and \textbf{BERTScore (F1)} \citep{bertscore_ref} to assess semantic similarity. All experiments were conducted using the top-$k=5$ retrieved context. Detailed generation prompts are provided in Appendix \ref{sec:prompts}. The comparative results are summarized in Table \ref{tab:generation_performance} and visualized in Figure \ref{fig:generation_res}.

\begin{table*}[!t]
    \centering
    \small 
    \caption{Generation Performance Comparison across FinanceBench, Mini-Wiki, and SeaCompany ($k=5$). Best results are \textbf{bolded}, and best baseline results are \underline{underlined}.}
    \label{tab:generation_performance}
    \begin{tabular*}{\textwidth}{@{\extracolsep{\fill}}l cc cc cc}
        \toprule
        & \multicolumn{2}{c}{\textbf{FinanceBench}} & \multicolumn{2}{c}{\textbf{Mini-Wiki}} & \multicolumn{2}{c}{\textbf{SeaCompany}} \\
        \cmidrule(lr){2-3} \cmidrule(lr){4-5} \cmidrule(lr){6-7}
        \textbf{Model} & BERT (F1) & ROUGE-L & BERT (F1) & ROUGE-L & BERT (F1) & ROUGE-L \\
        \midrule
        VSS & 0.8724 & 0.1689 & 0.8987 & 0.4704 & 0.8858 & 0.2760 \\
        VSS w/ Re-ranker & 0.8707 & 0.1739 & \underline{0.9053} & 0.4874 & 0.8867 & 0.2846 \\
        Sparse & 0.8513 & 0.1586 & 0.8991 & 0.4818 & 0.8895 & 0.2860 \\
        Hybrid & 0.8690 & 0.1627 & 0.8944 & 0.4790 & \underline{0.8934} & \underline{0.3087} \\
        \midrule
        ReAct & 0.7466 & 0.0195 & 0.9008 & \underline{0.5311} & 0.8860 & 0.2799 \\
        DeepSieve & 0.8380 & 0.1081 & 0.9005 & 0.4973 & 0.8522 & 0.2211 \\
        RAPTOR & \underline{0.8741} & \underline{0.1919} & 0.8979 & 0.5100 & 0.8847 & 0.2688 \\
        \midrule
        \textbf{Orion-RAG (Ours)} & \textbf{0.8822} & \textbf{0.2156} & \textbf{0.9119} & \textbf{0.5871} & \textbf{0.8955} & \textbf{0.3212} \\
        \midrule
        \textit{Rel. Improv.} & \textit{+0.93\%} & \textit{+12.35\%} & \textit{+0.73\%} & \textit{+10.54\%} & \textit{+0.24\%} & \textit{+4.05\%} \\
        \bottomrule
    \end{tabular*}
\end{table*}

\noindent \textbf{Results.}  Table \ref{tab:generation_performance} shows that Orion-RAG consistently achieves state-of-the-art performance across all benchmarks. Most notably, on the fact-intensive FinanceBench, our method secured a 12.35\% relative improvement in ROUGE-L over RAPTOR, confirming that superior retrieval precision translates directly into high-fidelity generation. Similarly, on Mini-Wiki, Orion-RAG led with a 10.54\% improvement, outperforming iterative models like ReAct which—while competitive on general knowledge—suffered severe breakdowns on complex tasks due to error cascading. Even on the fragmented SeaCompany dataset, where lexical matching is critical, Orion-RAG outperformed the strong Hybrid baseline by 4.05\%, demonstrating robust adaptability across diverse data topologies.


\noindent \textbf{Impact of Context Length \& Components.} We also investigated how text segmentation granularity affects the generator. In contrast to retrieval, which benefits from finer granularity, our ablation study on Mini-Wiki reveals that generation quality peaks at larger context windows (2000 characters). This highlights a structural dichotomy: retrieval requires precision to minimize noise, while generation requires breadth for narrative coherence and to capture sufficient background for complex reasoning (see Appendix \ref{subsec:ablation_generation}). Furthermore, we analyze the specific contributions of \textit{Query Expansion} and \textit{Document Pruning} in Appendix \ref{subsec:ablation_components}, verifying that while expansion is universally critical for structural alignment, pruning effectively reduces semantic noise to mitigate potential hallucinations in narrative-heavy domains. 



\subsection{Runtime Efficiency Analysis}
\label{subsec:runtime}

Beyond quality metrics, practical deployment requires efficiency. We measured the total runtime on the SeaCompany dataset (8 concurrent requests), strictly distinguishing between Offline Index Construction and Online Retrieval.

As shown in Table \ref{tab:runtime_comparison}, while ReAct and DeepSieve require minimal indexing time, their inference latency is prohibitive (over 800s) due to iterative reasoning loops. In contrast, Orion-RAG invests 129.57s in offline tagging but achieves a blazing fast online retrieval of 99.68s—an 8.6$\times$ speedup over ReAct. Even compared to RAPTOR, which also uses pre-computed structures, Orion-RAG is significantly faster in both construction and retrieval, proving its viability for high-concurrency production environments. 

\subsection{Human-in-the-Loop Optimization}
\label{subsec:hitl}

A\hspace{1ex}unique\hspace{1ex}advantage\hspace{1ex}of\hspace{1ex}Orion-RAG is its interpretability: unlike opaque dense vectors, our explicit Tag Index allow for precise human intervention. To validate this, we analyzed a failure case where the automated tagger missed a query-specific concept.

As detailed in Figure \ref{fig:hitl_case}, the query specifically inquired about BDO Unibank's ``diversified business model.'' The initial automated tags correctly identified general banking concepts (e.g., \textit{universal banking}, \textit{financial firm}) but missed the specific phrasing of the business model. Consequently, the dense retriever assigned a suboptimal L2 distance of $0.5744$.

By injecting the specific tag ``diversified business model'' into the document's path—a simple human-in-the-loop action—the semantic distance was reduced to $0.4528$ (a $\sim$21\% improvement in proximity). 
This allows experts to fine-tune retrieval by directly refining tags.

\begin{figure}[!t]
    \centering
\begin{tcolorbox}[
        colback=gray!5!white,
        colframe=black!75!white,
        title=\textbf{Case Study: Semantic Injection via Human Feedback},
        fonttitle=\bfseries\small,
        boxrule=0.8pt,
        arc=2pt,
        left=2pt, right=2pt, top=2pt, bottom=2pt
    ]
    \scriptsize\ttfamily
    \textbf{1. Query Request:} \\
    "How does BDO Unibank's \underline{diversified business model} help it maintain resilience during economic cycles?"
    
    \vspace{0.2cm}
    \textbf{2. Target Document (Gold):} \\
    \textbf{Ticker:} BDO (BDO Unibank) \\
    \textbf{Content Preview:} "The firm operates as a full-service universal bank, offering a comprehensive suite..."
    
    \vspace{0.2cm}
    \textbf{3. Pre-Intervention State:} \\
    \textbf{Current Tags:} ['universal banking services', 'financial firm', 'economy', 'BDO Unibank', 'SM Group', 'Digital banking', ...] \\
    \textbf{Metric (L2 Distance):} 0.5744 \\
    \textbf{Status:} \textcolor{red}{Suboptimal Match (Rank > 5)}
    
    \vspace{0.1cm}
    \hrule
    \vspace{0.1cm}
    
    \textbf{4. Human Intervention:} \\
    \textbf{Action:} Inject Tag ["diversified business model"]
    
    \vspace{0.1cm}
    \hrule
    \vspace{0.1cm}
    
    \textbf{5. Post-Intervention State:} \\
    \textbf{Refined Tags:} [..., 'universal banking services', \textbf{'diversified business model'}] \\
    \textbf{Metric (L2 Distance):} \textbf{0.4528} (\textcolor{green!60!black}{$\downarrow$ 0.1216 Improved}) \\
    \textbf{Status:} \textcolor{green!60!black}{Top-3 Retrieval Secured}
    \end{tcolorbox}
    \caption{HITL refinement: Injecting a specific tag reduces semantic distance, securing correct retrieval.}
    \label{fig:hitl_case}
\end{figure}

\begin{table}[t!]
    \centering
    \caption{Runtime on SeaCompany (8 concurrent).}
    \label{tab:runtime_comparison}
    \resizebox{\columnwidth}{!}{
        \begin{tabular}{lcccc}
            \toprule
            \textbf{Method} & \textbf{Index Const. (s)} & \textbf{Retrieval (s)} & \textbf{Total (s)} \\
            \midrule
            \textbf{Orion-RAG} & 129.57 & \textbf{99.68} & \textbf{229.25} \\
            RAPTOR & 164.71 & 232.48 & 397.19 \\
            ReAct-RAG & $\approx$ 5.00 & 858.14 & 863.14 \\
            DeepSieve & $\approx$ 5.00 & 1182.17 & 1187.17 \\
            \bottomrule
        \end{tabular}
    }
\end{table}
\FloatBarrier
\section{Conclusion}
\label{sec:conclusion}
In this paper, we introduced Orion-RAG, a framework tailored for retrieving information from fragmented, semi-structured enterprise data. By automating the extraction of hierarchical ``paths'' (Master Tags $\rightarrow$ Paragraph Tags), Orion-RAG structures raw documents locally without requiring global knowledge graphs. Our experiments demonstrate that this approach significantly outperforms complex iterative agents like ReAct in noisy, real-world environments, offering a robust balance between retrieval accuracy and deployment simplicity. Future work will focus on expanding evaluations to specialized domains such as legal and medical workflows, as well as conducting rigorous cost-latency benchmarking against token-heavy agentic baselines.



\section*{Limitations}
While Orion-RAG demonstrates robust performance across diverse datasets, it still has some limitations: (1) Its effectiveness relies heavily on the presence of extractable entities within the source text. In scenarios where paragraphs are highly abstract or lack distinct subjects, the generated paths may become generic or sparse, potentially reducing the efficiency of the path-based index. (2) The framework is sensitive to hyperparameter configurations, such as chunk sizes and pruning thresholds. As observed in our ablation studies, suboptimal settings can lead to either aggressive signal loss or excessive noise accumulation, suggesting that domain-specific tuning is often necessary for optimal results. (3) The performance of the rewriting and path generation is contingent on prompt design. Minor variations in instruction prompts can alter the granularity of generated tags, which may affect downstream retrieval precision in specialized contexts. Although these limitations highlight current constraints, they also point to valuable directions for future optimization and robustness research.

\FloatBarrier
\bibliography{acl_latex}

\clearpage
\appendix

\section{Appendix: Benchmarks}
\label{sec:appendix_benchmarks}

\subsection{Datasets and Pre-processing}
\label{subsec:datasets}

To evaluate the performance of our proposed system, Orion-RAG, particularly in realistic scenarios involving fragmented and graphless semi-structured data, we selected three distinct benchmarks: FinanceBench \citep{financebench_ref}, Mini-Wiki \citep{mini_wiki_ref}, and a custom dataset SeaCompany. These datasets were chosen because they exemplify ``graphless'' environments—while implicit connections exist via domains, entities, or themes, the corpora lack uniform schemas or pre-existing knowledge graphs. To strictly simulate real-world retrieval challenges where documents are long and unstructured, we applied specific processing strategies to each dataset before ingesting them into our Data Augmentation System (Section \ref{subsec:augmentation}).

\paragraph{FinanceBench.}
This dataset represents a domain-specific challenge, consisting of 150 QA pairs derived from real-world commercial financial reports. Since the original evidence texts are long-form excerpts (often exceeding 10,000 characters), we treated each as a standalone commercial document to mimic an enterprise application environment. We utilized the complete set of 150 QA pairs and their corresponding documents. These documents were processed by our augmentation pipeline to generate Master Tags and segment-level Path Tags.

\paragraph{Mini-Wiki.}
To assess performance on general knowledge, we utilized the Mini-Wiki dataset, which contains 3,200 documents and 918 QA pairs. In its raw form, Mini-Wiki provides short evidence passages. Treating these isolated snippets as independent documents would trivialize the retrieval task and fail to reflect the complexity of real-world corpora. To reconstruct realistic document granularity, we adopted a concatenation strategy: distinct evidence passages were merged into groups of 75 to form larger ``synthetic documents,'' mimicking the semantic density and length of complete Wikipedia articles. For evaluation efficiency, we randomly sampled 100 QA pairs from the original set. The synthetic corpus underwent the same augmentation and indexing process as FinanceBench, ensuring that retrieval was performed against a dense, large-scale backdrop rather than pre-segmented snippets.

\paragraph{SeaCompany.}
To specifically evaluate retrieval capabilities on highly fragmented and discrete data, we constructed SeaCompany, a manually collected dataset comprising profiles of 102 Southeast Asian corporations. The corpus covers diverse attributes such as financial revenue, business scope, and foundational company information. Crucially, the dataset is deliberately designed to be fragmented: each document section focuses strictly on a single aspect of a company without providing a holistic overview, effectively creating independent ``islands'' of information. This structure serves as a rigorous testbed for the system's ability to integrate scattered knowledge. To generate high-quality evaluation data, we utilized the RAGen framework \citep{tian2025ragen}, a domain-specific data generation tool, to synthesize 200 Question-Answer (QA) pairs tailored to this corpus. Each company profile was input as an independent document and segmented by paragraph for indexing.

\section{Appendix: Reproducibility}
\label{sec:reproducibility}

\subsection{Experimental Details}
\label{subsec:exp_details}

All experiments were conducted on a workstation equipped with four NVIDIA RTX 3090 GPUs (24GB VRAM). The framework was implemented in Python, and we initialized all stochastic components with a fixed random seed (42) to ensure reproducibility. To guarantee a fair comparison across all methods, we utilized Milvus \citep{milvus_ref} as the unified vector database for index management and storage. Similarly, bge-large-en-v1.5 \citep{bge_ref} served as the consistent embedding model for encoding queries and document chunks across both Orion-RAG and all baseline approaches.

\subsubsection{Orion-RAG Implementation}
\label{subsubsec:orion_impl}

\paragraph{Model Architectures.}
We employed a tiered model strategy to balance performance and computational cost. The Data Augmentation System utilized GPT-4o \citep{gpt4o_ref} for high-quality structure induction, specifically for generating Master and Paragraph Tags. For the online Retrieval and Generation System, we utilized GPT-4o-mini to optimize inference latency and throughput.

\paragraph{Chunking and Retrieval Settings.}
We adopted fixed-character splitting strategies tailored to dataset characteristics: 500 characters for FinanceBench to isolate specific figures, and 2000 characters for Mini-Wiki to preserve narrative context. The retrieval pipeline consists of two stages. In the coarse filtering stage, we retrieve the top-$3k$ candidates via the Tag Index and an auxiliary $\frac{2}{5}k$ candidates via the Sparse Index. In the re-ranking stage, candidates are fused using Reciprocal Rank Fusion (RRF) with weights set to $\alpha=0.25$ (Tag Score), $\beta=0.25$ (Semantic Score), and $\gamma=0.5$ (Sparse Score), with a fusion constant $c=60$. The final number of retrieved documents per subquery is fixed at $k=5$.

\subsubsection{Baseline Implementation}
\label{subsubsec:baselines}

To provide a rigorous comparison, we evaluated Orion-RAG against several representative retrieval-augmented generation systems. Implementation details were meticulously matched to ensure fairness: unless otherwise stated, all baselines operated on the same Milvus-based vector store and utilized bge-large-en-v1.5 for embeddings. Furthermore, to strictly isolate the contribution of the retrieval architecture, most baselines (specifically Dense Retrieval, Sparse Retrieval, Hybrid Retrieval, Re-ranker, and RAPTOR) reused the exact same generator module (GPT-4o-mini) as Orion-RAG. The notable exception is DeepSieve, which employs a specialized, integrated reasoning framework where the generation process is coupled with its iterative logic.

\paragraph{Vector Semantic Search (VSS by Dense Retireval)}
This baseline represents the standard semantic search approach. It retrieves the top-$k=5$ chunks based on L2 distance (cosine similarity) from the dense full-text index. It is evaluated on both retrieval and generation metrics.

\paragraph{Sparse Retrieval}
This baseline employs classical lexical matching using the BM25 algorithm. It retrieves the top-$k=5$ chunks solely based on keyword overlap statistics. Like D-RAG, it contributes to both retrieval and generation benchmarks.

\paragraph{Standard Hybrid Retrieval}
This method combines dense and sparse signals via a single-stage fusion. It utilizes Reciprocal Rank Fusion (RRF) with equal weights (50\% Dense, 50\% Sparse) to merge results from the bge-large-en-v1.5 and BM25 indices, returning the top-$k=5$ documents for both retrieval and generation evaluation.

\paragraph{Dense Retrieval with Re-ranker}
An enhancement of the D-RAG baseline, this system first retrieves a candidate pool (Top-$N$, where $N>k$) using dense embeddings, followed by a post-retrieval re-ranking step. A GPT-4o-mini agent selects the final $k=5$ most relevant chunks. This baseline is evaluated only on generation metrics, as the re-ranking step optimizes context quality for the answer rather than raw retrieval index performance.

\paragraph{ReAct Self-Reflective RAG (ReAct-RAG)}
We implemented the multi-step reasoning strategy from \citet{react_ref}. The agent utilizes GPT-4o-mini for iterative reasoning and query expansion, with the retrieval tool configured to search the standard dense index (max 3 lookups). Following the open-source implementation\footnote{\url{https://github.com/ysymyth/ReAct}}, this baseline participates in both evaluation tracks using its cumulative retrieval outputs.

\paragraph{DeepSieve Iterative RAG}
Based on the framework by \citet{deepsieve_ref}, this baseline uses a structured approach for knowledge aggregation. We adapted the official implementation\footnote{\url{https://github.com/MinghoKwok/DeepSieve}} to use our standardized embedding model and generator, ensuring that performance differences reflect the agentic architecture rather than encoder quality. It involves multiple retrieval steps and is evaluated on both retrieval and generation.

\paragraph{RAPTOR}
We utilized the official implementation of Recursive Abstractive Processing \citep{raptor_ref}\footnote{\url{https://github.com/parthsarthi03/raptor}}, which constructs a hierarchical tree of text summaries. To ensure strict fairness, we modified the codebase to use bge-large-en-v1.5 embeddings and GPT-4o-mini for summarization. During inference, it traverses the tree to retrieve context (Top-$k=5$). Since RAPTOR retrieves synthesized summaries rather than original ground truth text chunks, standard hit-rate metrics are inapplicable; thus, it is evaluated exclusively on generation performance.

\subsection{Evaluation Metrics}
\label{subsec:eval_metrics}

We evaluated our system comprehensively, assessing both the quality of the final generated answers and the precision of the intermediate retrieval module.

\paragraph{Generation Performance}
To assess generation quality, we employed two widely recognized metrics applied across all three datasets (\textbf{FinanceBench}, \textbf{Mini-Wiki}, and \textbf{SeaCompany}):
\begin{itemize}[nosep]
    \item \textbf{ROUGE-L} \citep{rouge_ref}: Measures the longest common subsequence between the generated response and the reference answer. It captures structural fidelity and fluency alignment.
    \item \textbf{BERTScore (F1)} \citep{bertscore_ref}: Utilizes contextual embeddings to evaluate semantic similarity. This metric provides a more robust assessment of meaning preservation than lexical overlap alone, correlating better with human judgment.
\end{itemize}

\paragraph{Retrieval Performance}
For retrieval evaluation, we focused exclusively on \textbf{FinanceBench} and \textbf{SeaCompany}. We excluded Mini-Wiki from this specific assessment due to the ambiguity introduced by its synthetic concatenation process, which makes precise chunk-level attribution difficult.

\noindent \textit{Ground Truth Definition:} Both FinanceBench and SeaCompany provide explicit mappings where each query corresponds to a specific source evidence document (or corporate profile). Since our data augmentation system segments these source documents into multiple chunks, we define the ground truth set $G_q$ for a query $q$ as: \textit{all sub-chunks derived from the specific source document associated with the query.}

We assessed retrieval performance using two metrics across $k \in \{3, 5, 10\}$:
\begin{itemize}[nosep]
    \item \textbf{Hit Rate@k:} Measures the percentage of queries where at least one relevant ground truth chunk is present in the top-$k$ retrieved results. A high Hit Rate indicates the system's ability to successfully locate correct information boundaries.
    \item \textbf{Precision:} Measures the proportion of relevant chunks within the retrieved set, defined as $|Retrieved_k \cap G_q| / k$. This metric is critical for evaluating multi-query or expanded retrieval strategies. While retrieving a large volume of documents can artificially inflate Hit Rate, low precision implies a high number of irrelevant ``distractor'' documents, which can introduce noise and hallucination in the downstream generator.
\end{itemize}

\subsection{Prompts}
\label{sec:prompts}

We provide the core prompt templates used in Orion-RAG. To ensure applicability across different benchmarks (FinanceBench, Mini-Wiki, SeaCompany), specific domain keywords are abstracted as \texttt{\{\{DOMAIN\}\}}.

\subsubsection{Data Augmentation Prompts}

The prompt for Paragraph Tag Generation is:

\begin{tcolorbox}[colback=bggray, colframe=black!70, arc=4mm, boxrule=1pt, left=6pt, right=6pt, top=6pt, bottom=6pt]
\small
\textbf{System:} You are a precise \texttt{\{\{DOMAIN\}\}} paragraph tag extractor. Return ONLY a JSON array of 2-3 short tags, no extra text. Each tag must be 1–4 words, contain no punctuation or quotes, and use canonical English forms.
Output order and meaning:
1) gist: a short title-like summary of the paragraph content
2) subjects: the main entities/subjects involved
3) domain: the most specific domain/field (e.g., geography, history, economy, culture, politics, demographics, events, science)
Use singular nouns where reasonable; capitalize proper nouns; deduplicate similar tags; avoid vague terms and numbers-only.

\textbf{User:} From the paragraph below, extract EXACTLY 3 tags in this order:
1) gist (summary)
2) subjects (entities/actors)
3) domain (specific field)
Return ONLY a JSON array of strings.

Paragraph:
\texttt{\{text\}}
\end{tcolorbox}

\vspace{0.5cm}
The prompt for Master Tag Generation is:

\begin{tcolorbox}[colback=bggray, colframe=black!70, arc=4mm, boxrule=1pt, left=6pt, right=6pt, top=6pt, bottom=6pt]
\small
\textbf{System:} You are a precise \texttt{\{\{DOMAIN\}\}} document tag extractor. Return ONLY a JSON array of strings, no extra text. Output at most the requested number of tags.
Each tag should:
- be 1–4 words; avoid punctuation; prefer canonical English surface forms (Wikipedia title style)
- include the main subject(s)/entities, geographic scope, central time period/era when relevant, major subtopics (history, politics, economy, culture, demographics, notable events), and important organizations/people mentioned frequently
- deduplicate/merge similar tags; avoid near-synonyms; prefer singular forms; avoid boilerplate like "Overview" or vague terms like "various".

\textbf{User:} From the article text below, extract up to \texttt{\{max\_tags\}} concise tags that best describe the article's subjects, geographic/temporal scope, and major subtopics. Return ONLY a JSON array of strings.

Text:
\texttt{\{text\}}
\end{tcolorbox}

\subsubsection{Retrieval and Generation Prompts}

The prompt for Query Rewritter is:

\begin{tcolorbox}[colback=bggray, colframe=black!70, arc=4mm, boxrule=1pt, left=6pt, right=6pt, top=6pt, bottom=6pt]
\small
\textbf{System:} You are a query expansion assistant for \texttt{\{\{DOMAIN\}\}} retrieval. Return ONLY a JSON array of strings. Your goals: (1) analyze complex user queries and identify multiple different search intents; (2) generate separate, short retrieval queries for different domains/topics; (3) decide quantity by complexity; (4) include specific entity names when intent involves specific targets; (5) expand regional groups (e.g., \texttt{\{\{REGION\_GROUP\}\}}) into specific countries as separate queries; (6) prefer coarse-grained, noun-based keywords (e.g., organization, company, suppliers, policy); avoid adjectives/adverbs; avoid ambiguous words like 'demand', 'supply', 'comparison' unless making two-side queries; (7) if conversation history implies specific focus, reflect it.

\textbf{User:} Task: Decompose the user query into up to \texttt{\{max\_n\}} short, retrieval-friendly sub-queries.
Constraints: each sub-query $\le$ 12 words; avoid punctuation and stopwords where possible; one entity per query when relevant;
Return ONLY a JSON array of strings. No explanations.

User Query: \texttt{\{q\}}
Conversation Hints (optional): \texttt{\{hist\}}
\end{tcolorbox}

\vspace{0.5cm}
The prompt for Document Pruning is:

\begin{tcolorbox}[colback=bggray, colframe=black!70, arc=4mm, boxrule=1pt, left=6pt, right=6pt, top=6pt, bottom=6pt]
\small
\textbf{System:} You are a \texttt{\{\{DOMAIN\}\}} RAG pruning assistant. Return ONLY the pruned text (plain text), no explanation. Goal: analyze the retrieved text chunk together with a sub-query and the original user query; remove sentences/paragraphs that are irrelevant to the sub-query and unhelpful to answer the original query. Keep the remaining content order; keep essential numbers, entities, and line items; prefer cash-flow related items when applicable. If nothing relevant remains, return an empty string. Output must be plain text only.

\textbf{Instructions:} 
- Remove lines/paragraphs that are irrelevant to the sub-query or do not help answer the original query.
- Keep numeric facts (amounts, dates, units), financial line items (e.g., cash flow statement rows), companies, and policy/organization mentions that relate to the sub-query.
- Return ONLY the pruned text, no commentary, no code fence.
\end{tcolorbox}

\vspace{0.5cm}
The prompt for the Answer Generation is:

\begin{tcolorbox}[colback=bggray, colframe=black!70, arc=4mm, boxrule=1pt, left=6pt, right=6pt, top=6pt, bottom=6pt]
\small
\textbf{System:} You are a \texttt{\{\{DOMAIN\}\}} QA assistant. Answer strictly based on the provided context. Be concise and precise; keep figures/units exactly as shown in the context. If the answer is a factual value, output ONLY the fact without restating the question or adding extra text.

\textbf{Instructions:}
- Use only the Context to answer the Question.
- If the answer is a number, include units and year explicitly.
- If the answer is a factual value, output only the fact (no preface or restating).
- Return ONLY the final answer text.
\end{tcolorbox}
\subsubsection{Baseline Prompts}

To ensure reproducibility, we explicitly present the adapted reasoning aggregation prompt used for the DeepSieve baseline. Note that we only display the specific components modified to align with our output format; all other planning and reasoning prompts remain identical to the official open-source implementation.

\begin{tcolorbox}[colback=bggray, colframe=black!70, arc=4mm, boxrule=1pt, left=6pt, right=6pt, top=6pt, bottom=6pt]
\small
\textbf{System:} You are a \texttt{\{\{DOMAIN\}\}} QA assistant. Answer strictly based on the provided steps (Context). Be concise and precise; keep figures/units exactly as shown in the context. If the answer is a factual value, output ONLY the fact without restating the question or adding extra text.

\textbf{User:} 
Original Question: \texttt{\{original\_question\}}

Subquestion Reasoning Steps:
\textit{<The following block is repeated for each retrieval step>}
\texttt{\{subquery\_id\}: \{actual\_query\} $\to$ \{answer\}}
\texttt{Reason: \{reason\}}
\textit{<End of repeated block>}

Based on the above reasoning steps, what is the final answer to the original question?

Please respond in JSON format:
\{
  "answer": "final\_answer",
  "reason": "final\_reasoning"
\}
Only output valid JSON. Do not add any explanation or markdown code block markers.
\end{tcolorbox}

\section{Appendix: Experiments}
\label{sec:appendix_experiments}

\subsection{Ablation Study on Retrieval Module}
\label{subsec:ablation_retrieval}

In this section, we analyze the impact of text segmentation on the retrieval stage, specifically focusing on the trade-off between semantic context and signal noise.

\paragraph{Settings.}
Chunk size is a critical hyperparameter in dense retrieval: chunks must be large enough to capture semantic context but small enough to prevent signal dilution. To determine the optimal segmentation strategy for the precision-sensitive FinanceBench dataset, we conducted an ablation study evaluating Orion-RAG's retrieval performance at $k=5$ across four distinct granularities: 200, 500, 2000, and 5000 characters.

\paragraph{Results.}
As shown in Table \ref{tab:ablation_chunksize_retrieval}, distinct trends emerged regarding the trade-off between precision and recall. Small chunks (200 chars) yielded the highest Precision (0.284), aligning with the need to extract specific financial figures. However, the 500-character setting achieved the optimal balance, delivering the peak Hit Rate (0.920) while maintaining competitive precision. Notably, performance degraded significantly at larger sizes (2000+ chars), confirming that excessive context introduces ``semantic noise'' that confuses the dense retriever in fine-grained tasks.

\begin{table}[h]
    \centering
    \caption{Ablation study on the impact of chunk size on retrieval performance ($k=5$). The 500-character setting achieves the best balance between coverage (Hit Rate) and accuracy (Precision).}
    \label{tab:ablation_chunksize_retrieval}
    \small
    \begin{tabular}{lcc}
        \toprule
        \textbf{Chunk Size} & \textbf{Hit Rate} & \textbf{Precision} \\
        \midrule
        200 chars  & 0.913 & \textbf{0.284} \\
        500 chars  & \textbf{0.920} & 0.237 \\
        2000 chars & 0.873 & 0.190 \\
        5000 chars & 0.887 & 0.181 \\
        \bottomrule
    \end{tabular}
\end{table}

\subsection{Ablation Study on Generation Module}
\label{subsec:ablation_generation}

While the retrieval module prefers finer granularity, the generation module often requires different context lengths. Here we evaluate the Generator's performance under varying context constraints.

\paragraph{Settings.}
While retrieval benefits from fine-grained segmentation, generation often requires broader context to synthesize coherent narratives, particularly in general knowledge domains. We investigated this dichotomy by evaluating generation quality on the MiniWiki dataset across the same spectrum of chunk sizes (200 to 5000 characters). The goal was to identify the threshold where the Generator (LLM) has sufficient context to answer complex questions without being overwhelmed by irrelevant tokens.

\paragraph{Results.}
The results in Table \ref{tab:ablation_chunksize_gen} reveal a contrasting trend to the financial retrieval task: generation quality consistently improved with larger contexts up to a point. The 200-character chunks resulted in the lowest scores (ROUGE-L 0.4522), indicating that overly fragmented text disrupts the semantic continuity required for fluent answers. The system peaked at 2000 characters (ROUGE-L 0.5871), validating that for narrative-heavy tasks, larger chunks provide the necessary comprehensive context. Performance plateaued and slightly dipped at 5000 characters, suggesting diminishing returns where added noise begins to outweigh the benefit of additional context.

\begin{table}[h]
    \centering
    \caption{Ablation study on generation performance across different chunk sizes on the MiniWiki dataset. Unlike retrieval, generation quality benefits from larger contexts, peaking at 2000 characters before diminishing.}
    \label{tab:ablation_chunksize_gen}
    \resizebox{0.85\linewidth}{!}{
        \begin{tabular}{lcc}
            \toprule
            \textbf{Chunk Size} & \textbf{BERTScore F1} & \textbf{ROUGE-L} \\
            \midrule
            200 chars  & 0.8977 & 0.4522 \\
            500 chars  & 0.9089 & 0.5547 \\
            2000 chars & \textbf{0.9119} & \textbf{0.5871} \\
            5000 chars & 0.9083 & 0.5785 \\
            \bottomrule
        \end{tabular}
    }
\end{table}

\subsection{Ablation Study on Pipeline Components}
\label{subsec:ablation_components}

Finally, we investigate the contribution of two auxiliary modules: \textit{Query Expansion} (via the Rewriting Agent) and \textit{Document Pruning}. This ablation study aims to decouple the impact of structural alignment from noise filtering across three diverse datasets: FinanceBench, MiniWiki, and SeaCompany.

\paragraph{Settings.}
We evaluated four configurations: (1) the full \textbf{Orion-RAG} pipeline; (2) \textbf{w/o Pruning}, where retrieved chunks are fed directly to the generator; (3) \textbf{w/o Expansion}, where raw user queries are used for retrieval; and (4) \textbf{w/o Both}. This setup tests the hypothesis that query expansion is essential for mapping user intent to our path-based indices, while pruning serves as a context regulator for the LLM.

\paragraph{Results.}
The results in Table \ref{tab:ablation_components} demonstrate that \textbf{Query Expansion} is the dominant factor for performance. Removing this module caused substantial degradation, particularly in the FinanceBench dataset (ROUGE-L dropping from 0.2156 to 0.1656). This confirms that expansion acts as a crucial bridge, normalizing natural language queries into ``path-aligned'' keyword combinations that maximize the efficacy of our structural index.

The impact of the \textbf{Document Pruning} module proved to be more domain-dependent. In the narrative-heavy MiniWiki dataset, pruning significantly boosted performance (ROUGE-L 0.5871 vs. 0.5404 without pruning), validating its role in reducing ``semantic noise'' for the generator. However, in fragmented domains like SeaCompany, aggressive pruning occasionally discarded useful signals, suggesting that the optimal pruning threshold is highly sensitive to domain characteristics and chunk size. We identify the dynamic tuning of this module as a promising direction for future research.

\begin{table}[h]
    \centering
    \caption{Ablation results (ROUGE-L). Expansion is key for structure alignment, while Pruning helps narrative tasks (MiniWiki) but needs tuning for fragmented data.}
    \label{tab:ablation_components}
    \small 
    \begin{tabular*}{\columnwidth}{@{\extracolsep{\fill}}lccc}
        \toprule
        \textbf{Configuration} & \textbf{Finance} & \textbf{MiniWiki} & \textbf{SeaCo.} \\
        \midrule
        Orion-RAG (Full) & 0.2156 & \textbf{0.5871} & 0.3212 \\
        w/o Pruning & \textbf{0.2217} & 0.5404 & \textbf{0.3589} \\
        w/o Expansion & 0.1656 & 0.5811 & 0.2868 \\
        w/o Exp. \& Prun. & 0.1650 & 0.4784 & 0.3029 \\
        \bottomrule
    \end{tabular*}
\end{table}

\end{document}